\documentclass{esannV2}
\usepackage[dvips]{graphicx}
\usepackage[latin1]{inputenc}
\usepackage{amssymb,amsmath,array}
\usepackage{xcolor}
\usepackage{booktabs}
\usepackage{url}
\usepackage{multirow}
\usepackage{wrapfig}
\usepackage[sorting=none, minnames=1,maxnames=1]{biblatex}
\AtBeginBibliography{\footnotesize} 

\begin{document}
\title{Boundary-Constrained Diffusion Models for Floorplan Generation: Balancing Realism and Diversity}

\author{Leonardo Stoppani$^1$, Davide Bacciu$^1$ and Shahab Mokarizadeh$^2$
	%
	%
	\vspace{.3cm}\\
	%
	1- University of Pisa - Dept of Computer Science \\
	Largo Bruno Pontecorvo 3, Pisa, Italy
	%
	\vspace{.1cm}\\
	2- H\&M Group \\
	Stockholm, Stockholm County, Sweden\\
}

\maketitle

\sloppy

\begin{abstract} 
	Diffusion models have become widely popular for automated floorplan generation, producing highly realistic layouts conditioned on user-defined constraints. However, optimizing for perceptual metrics such as the Fre\'chet Inception Distance (FID) causes limited design diversity. To address this, we propose the Diversity Score (DS), a metric that quantifies layout diversity under fixed constraints. Moreover, to improve geometric consistency, we introduce a Boundary Cross-Attention (BCA) module that enables conditioning on building boundaries. Our experiments show that BCA significantly improves boundary adherence, while prolonged training drives diversity collapse undiagnosed by FID, revealing a critical trade-off between realism and diversity. Out-Of-Distribution evaluations further demonstrate the models' reliance on dataset priors, emphasizing the need for generative systems that explicitly balance fidelity, diversity, and generalization in architectural design tasks.
\end{abstract}

\section{Introduction}


The field of automated floorplan generation has seen a clear shift from early Deep Learning (DL) methods \cite{wu2019rplan, hu2020graph2plan} to more recent generative architectures \cite{gen_floorplan_survey}. Approaches based on Denoising Diffusion Probabilistic Models (DDPMs) \cite{shabani2023housediffusion, hong2024cons2plan} surpassed previous ones based on Generative Adversarial Networks (GANs) \cite{Nauata2021HouseGAN++, Nauata2020HouseGAN}, becoming the most popular approach for the creation of vectorized layouts from user-defined constraints.

The primary goal of recent research has been to generate layouts that are both geometrically valid and stylistically realistic. Consequently, the community has largely focused on optimizing for metrics like Fre\'chet Inception Distance (FID) \cite{fid}, which measures the perceptual similarity between generated samples and a ground-truth dataset. However, we argue that this intense focus on perceptual realism is hurting concrete progress in floorplan generation. Although a low FID score indicates realistic samples, it is not enough to ensure generated layouts satisfy architectural needs such as diversity and generalization to novel scenarios.

We identify three main limitations in current approaches: (i) they lack the ability to enforce hard spatial constraints, such as building boundaries; (ii) they do not ensure diversity of generated layouts for a given set of constraints; and (iii) they struggle to generalize to Out-Of-Distribution (OOD) scenarios, limiting their applicability in real-world architectural design tasks.

Our work, motivated by a concrete application need of a global fashion retailer, addresses these gaps through three main contributions. First, we introduce a dedicated boundary conditioning mechanism, the Boundary Cross-Attention (BCA) module. Second, we propose the Diversity Score (DS), a novel metric designed to quantify the diversity of generated layouts under the same conditioning. Finally, we conduct an OOD evaluation to assess the model's generalization capabilities beyond the training data distribution. Experimental results demonstrate that while our BCA mechanism improves boundary adherence, prolonged training leads to overfitting, as evidenced by a decline in the DS. The OOD tests further reveal that current models rely heavily on dataset-specific priors rather than robust geometric reasoning.



\section{Boundary-Constrained Vector Floorplan Generation}

Our model builds upon HouseDiffusion (HD) \cite{shabani2023housediffusion}, which defines a generative process for vectorized floorplan layouts using DDPMs. In the HD's framework, each floorplan is described as a set of polygonal loops $P={P_1,P_2,...,P_N}$, where $P_i={C_{i,1},C_{i,2},...,C_{i,N_i}|C_{i,j}\in\mathbb{R}^2}$ corresponds to a room and is defined by its corner coordinates. Originally, the model allows conditioning only on a user-defined bubble diagram represented as a graph adjacency matrix $G \in \mathbb{R}^{M \times 3} $. In our extension, we introduce an additional conditioning on a user-defined boundary, the polygonal loop $B = \{C_1,C_2,...,C_N\vert C_i \in \mathbb{R}^2\}$ that encloses the union of all room polygons.

\begin{figure}[!b]
	\centering
	\includegraphics[width=0.7\textwidth]{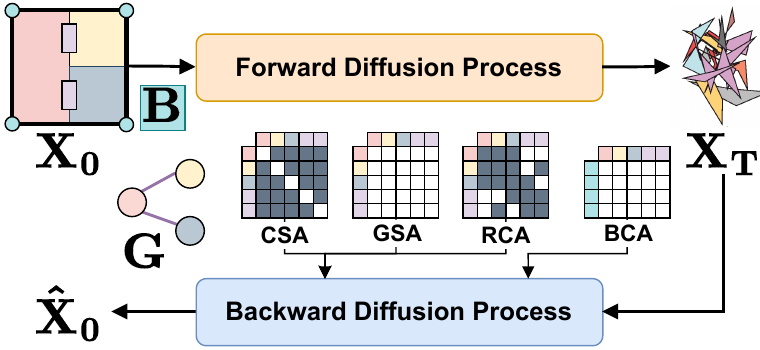}
	\caption{Our HouseDiffusion (HD) framework extended to include boundary conditioning.}\label{Fig:arch}
\end{figure}

Figure \ref{Fig:arch} illustrates HD's framework, where $X$ are the concatenated and padded room corner coordinates. The forward diffusion process gradually corrupts the initial corner coordinates $X_0$ over $T$ timesteps by adding Gaussian noise. The backward process aims to reverse this procedure to recover clean coordinates $X_0$ from pure noise $X_T$. HD trains a Transformer to predict the added noise component, given the noisy coordinates $X_T$ and the graph constraints $G$, which are implemented as three specialized attention masks: (i) Component-wise Self-Attention (CSA), which limits attention to corners within the same room or door; (ii) Global Self-Attention (GSA), a standard self-attention mechanism applied between every pair of corners across all rooms; and (iii) Relational Cross-Attention (RCA), which restricts attention between rooms and doors that are connected in the input constraint graph.

\paragraph{Boundary Cross-Attention Module.} Our primary modification to HD is the Boundary Cross-Attention (BCA) module. First, we apply self-attention to the embedding of the boundary corners $B$ to make each corner aware of the entire boundary shape. Then, we allow the room corner representations $X$ to attend to these enriched boundary embeddings with a standard cross-attention. We integrate the BCA output with the existing attention outputs with a simple summation. The BCA module allows us to add the boundary information directly as input to the model $\epsilon_{\theta}(X_t, t, G, B)$, which is trained via classifier-free guidance. We randomly mask the boundary with a probability $p_{\text{drop}}^B$ during training, which disables the BCA module and teaches the model to generate layouts both with and without the boundary condition. At inference, this allows us to steer the generation by blending the noise predictions from the boundary-conditional model and the boundary-unconditional model using a guidance scale $\lambda \in [0,1]$:
\[
	\tilde{\epsilon}_{\theta} \left( Z_t, t, G, B \right) =
	\lambda \, \epsilon_{\theta} \left(  Z_t, t, G, B \right) +
	(1 - \lambda) \, \epsilon_{\theta} \left(  Z_t, t, G, \varnothing \right).
\]
A larger $\lambda$ value more strictly enforces boundary adherence, while a smaller value promotes greater creative diversity in the generated layouts.

Our BCA mechanism differs slightly from the boundary conditioning strategy adopted in Cons2Plan \cite{hong2024cons2plan}. In Cons2Plan, boundary information is encoded using a convolutional feature extractor and integrated into the transformer through cross-attention between the CNN feature embeddings and the room corner tokens. In contrast, our BCA mechanism performs cross-attention directly between boundary corner tokens and room corner tokens. 

\paragraph{Diversity Score.} We introduce the Diversity Score (DS), a metric designed to quantify the variability of layouts generated from the same input conditions. Given a set of generated floorplan samples $\{x_i\}$ for a fixed input condition, the DS is computed as the trace of the covariance matrix of their features:
\[
	\text{DS} = \text{Tr}(\text{Cov}(\{f(x_i)\})).
\]
Here, $f$ is the feature extraction function from a pretrained InceptionV3 model \cite{Szegedy_2016_CVPR}. Unlike FID, which measures similarity to a reference dataset, DS directly quantifies the model's ability to explore different valid layouts for the same input constraints. A high DS indicates a creative model capable of producing diverse options, while a low DS suggests overfitting and mode collapse, where the model generates nearly identical outputs for a given condition.

\section{Experiments}

\paragraph{Experimental Setup.} Our model is implemented as an extension of the official HD PyTorch code\footnote{\url{https://github.com/aminshabani/house_diffusion}}. We trained the model for 400,000 steps with a batch size of 400 on a single NVIDIA L4 GPU. The Adam optimizer was used with an exponentially decaying learning rate from 1e-3 to 1e-5. The diffusion process uses 1000 steps and a cosine noise schedule. We choose a $p_{\text{drop}}^B$ of 0.1 during training. Following HD, we set the discrete loss timestep threshold to 32 and sample the number of corners for each room from a histogram of the training data. For evaluation, FID, Graph Compatibility (GC) \cite{shabani2023housediffusion} and Boundary Compatibility (BC) \cite{hong2024cons2plan} scores are computed over 512 generated samples, while for DS we generate 100 samples for four distinct input conditions and report the mean score. For our OOD analysis, first we pretrain the model on RPLAN, then we fine-tune it on two smaller, specialized datasets with a varying number of examples (shots).

\paragraph{Model Performance Comparison.} We first evaluate our model on RPLAN \cite{wu2019rplan}, a public dataset of 80,000 residential floorplans\footnote{The dataset is available for download at \url{http://staff.ustc.edu.cn/~fuxm/projects/DeepLayout/index.html}.}. Table \ref{tab:comparison} shows that our method achieves the lowest BC score (0.04), improving upon existing approaches while maintaining competitive FID and GC scores. This suggests that our BCA mechanism effectively enforces boundary adherence, rather than simply learning a trivial mapping from input boundaries to training layouts.

\begin{table}[!htb]
	\centering
	\renewcommand{\arraystretch}{1.0}
	\setlength{\tabcolsep}{4pt}
	\begin{tabular}{c|c|c|c}
		\toprule
		Model                               & FID (\textdownarrow)   & GC (\textdownarrow)    & BC (\textdownarrow)    \\
		\midrule
		HD \cite{shabani2023housediffusion} & 10.35$\pm$0.23         & 1.73$\pm$0.00          & \textemdash            \\
		Graph2Plan \cite{hu2020graph2plan}  & 30.48$\pm$0.00         & \textbf{1.45$\pm$0.25} & 0.11$\pm$0.00          \\
		Cons2Plan \cite{hong2024cons2plan}  & \textbf{6.80$\pm$0.13} & 1.50$\pm$0.08          & 0.06$\pm$0.00          \\
		Ours                                & 10.74$\pm$0.00         & 1.73$\pm$1.41          & \textbf{0.04$\pm$0.02} \\
		\bottomrule
	\end{tabular}
	\caption{Comparison of FID, GC, and BC on 512 floorplans. The best result for each metric is highlighted in bold.} \label{tab:comparison}
\end{table}


However, to investigate the consequences of prolonged training, we analyze performance at different checkpoints using our proposed DS. As shown in Table \ref{tab:training_results}, while FID and BC scores improve with more training epochs, the DS collapses from a peak of 52.69 down to 33.59. This inverse correlation reveals a critical trade-off between realism and diversity. It demonstrates that naively optimizing for standard realism metrics is counterproductive, as it pushes the model towards overfitting and mode collapse, causing it to lose the ability to generate a diverse set of novel solutions for a given input condition.

\begin{table}[!htb]
	\centering
	\renewcommand{\arraystretch}{1.0}
	\setlength{\tabcolsep}{4pt}
	\begin{tabular}{c|c|c|c|c}
		\toprule
		Steps ($10^3$) & FID (\textdownarrow) & GC (\textdownarrow)    & BC (\textdownarrow)     & DS (\textuparrow)       \\
		\midrule
		400            & 10.74                & \textbf{1.73$\pm$1.42} & \textbf{0.04$\pm$0.028} & 33.59$\pm$10.10         \\
		300            & \textbf{10.71}       & 1.88$\pm$1.44          & 0.04$\pm$0.028          & 35.49$\pm$8.94          \\
		200            & 10.77                & 1.89$\pm$1.42          & 0.05$\pm$0.028          & 34.28$\pm$8.84          \\
		100            & 11.38                & 2.10$\pm$1.48          & 0.06$\pm$0.032          & 37.84$\pm$5.75          \\
		50             & 14.14                & 2.57$\pm$1.73          & 0.08$\pm$0.033          & 40.27$\pm$5.38          \\
		30             & 16.33                & 3.17$\pm$1.71          & 0.10$\pm$0.035          & 42.95$\pm$4.56          \\
		20             & 18.26                & 3.41$\pm$1.83          & 0.11$\pm$0.034          & 48.40$\pm$6.76          \\
		10             & 25.98                & 3.95$\pm$1.77          & 1.00$\pm$0.000          & \textbf{52.69$\pm$5.91} \\
		\bottomrule
	\end{tabular}
	\caption{Comparison of FID, GC, BC and DS across different training steps. The best result for each metric is highlighted in bold.}
	\label{tab:training_results}
\end{table}

\begin{figure}[!htb]
	\centering
	\includegraphics[width=0.8\textwidth]{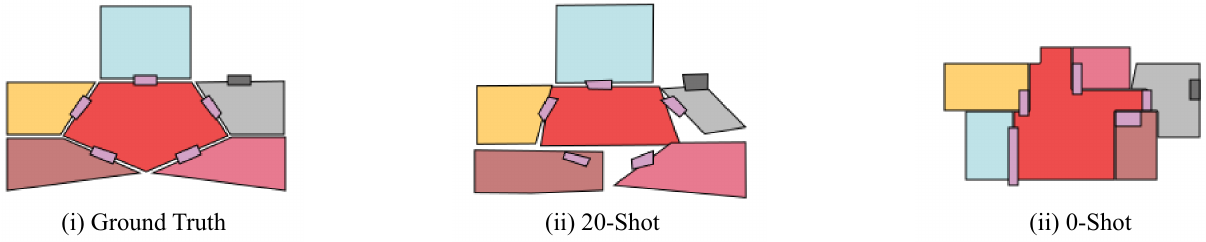}
	\caption{Example of failure due to the model's reliance on the training data distribution for the number of corners of each room.}\label{Fig:ood-samples}
\end{figure}

\paragraph{OOD Evaluation.} To further probe the model's generalization capabilities, we evaluate its few-shot performance on two Out-of-Distribution (OOD) datasets. The results are detailed in Table \ref{tab:few-shot-results}. On the drift dataset, representing a small distribution shift created by swapping the living room and balcony types in RPLAN, the model adapts effectively, showing substantial FID improvement with only a few fine-tuning shots. The synthetic dataset, which represents a large shift featuring 20 procedurally generated floorplans with a novel architectural style, proves far more challenging. As shown in Figure \ref{Fig:ood-samples}, we choose these floorplans to have a central pentagonal room connected to five surrounding rooms, introducing diagonal walls and more complex geometries not seen during training.

\begin{table}[!htb]
	\centering
	\renewcommand{\arraystretch}{1.0}
	\setlength{\tabcolsep}{4pt}
	\begin{tabular}{c|cc|cc}
		\toprule
		      & \multicolumn{2}{c|}{Drift Dataset} & \multicolumn{2}{c}{Synthetic Dataset}                                                 \\
		Shots & FID (\textdownarrow)               & GC (\textdownarrow)                   & FID (\textdownarrow) & GC (\textdownarrow)    \\
		\midrule
		0     & 38.69                              & 4.16$\pm$2.34                         & 290.00               & 2.59$\pm$2.00          \\
		1     & 31.35                              & 3.06$\pm$2.02                         & \textemdash          & \textemdash            \\
		2     & 44.38                              & \textbf{1.60$\pm$1.45}                & \textemdash          & \textemdash            \\
		4     & 29.50                              & 2.01$\pm$2.08                         & \textemdash          & \textemdash            \\
		5     & 31.13                              & 2.01$\pm$1.57                         & 182.06               & \textbf{1.74$\pm$1.03} \\
		10    & 29.96                              & 1.99$\pm$1.84                         & 150.35               & 1.74$\pm$1.16          \\
		20    & \textbf{26.47}                     & 1.89$\pm$1.62                         & \textbf{121.54}      & 1.80$\pm$1.07          \\
		\bottomrule
	\end{tabular}
	\caption{Few-shot fine-tuning results on drift and synthetic datasets. The best result for each metric is highlighted in bold.} \label{tab:few-shot-results}
\end{table}

The OOD experiments reveal the model's reliance on training data priors, which limits its ability to generalize to novel configurations. While it adapts well to the minor shift in the drift dataset, it struggles significantly with the synthetic dataset. As shown in Figure \ref{Fig:ood-samples}, the model fails to correctly infer that the central room must be pentagonal to accommodate five adjacent rooms. This failure is due to its mechanism for determining room complexity: it samples the number of corners for each room from a histogram derived from the training data. This heuristic works for RPLAN, where room shapes follow a consistent distribution. However, the model's inability to infer geometric complexity directly from the graph context results in unrealistic layouts when encountering unfamiliar architectural styles.

\section{Conclusions}
In this work, we addressed key limitations in automated floorplan generation. We introduced the BCA module to enforce hard spatial constraints, achieving state-of-the-art boundary adherence. We proposed the Diversity Score (DS), a metric that quantifies diversity under fixed conditions and exposes the inherent trade-off between realism and diversity. Furthermore, our OOD experiments revealed a strong dependence on training data priors, limiting generalization to novel spatial configurations. Overall, these findings highlight the need to move beyond purely data-driven paradigms toward more principled generative frameworks based on reasoning, effectively paving the way for models that better balance realism, diversity, and generalization in architectural design.


\printbibliography


\end{document}